\documentclass[letterpaper, 10pt, conference]{ieeeconf}  

\IEEEoverridecommandlockouts                              

\usepackage{amsmath}
\usepackage{amssymb}
\usepackage[hidelinks]{hyperref}
\usepackage{color}
\usepackage{graphicx}
\usepackage{subcaption}
\usepackage{multirow}
\usepackage{booktabs}
\usepackage{derivative}

\usepackage{algorithm}
\usepackage{algpseudocode}
\usepackage{flushend}

\DeclareMathOperator*{\argmin}{arg\,min}

\newcommand{\ph}[1]{{\textbf{#1}.}} 

\title{\LARGE \bf
Safe and Efficient Navigation in Extreme Environments using\\ Semantic Belief Graphs
}

\author{Muhammad Fadhil Ginting$^{1}$, Sung-Kyun Kim$^{2}$, Oriana Peltzer$^{1}$, Joshua Ott$^{1}$, Sunggoo Jung$^{2}$, \\ 
Mykel J. Kochenderfer$^{1}$, and Ali-akbar Agha-mohammadi$^{2}$ %
\thanks{$^{1}$Department of Aeronautics \& Astronautics, Stanford University, Stanford, CA, USA
        {\tt\small {\{ginting, peltzer, joshuaott, mykel\}}@stanford.edu}}%
\thanks{$^{2}$NASA Jet Propulsion Laboratory, California Institute of Technology, Pasadena, CA, USA
        {\tt\small {\{sung.kim, sunggoo.jung, aliagha\}}@jpl.nasa.gov}}%
}

\begin{document}

\maketitle

\begin{abstract}
To achieve autonomy in unknown and unstructured environments, 
we propose a method for semantic-based planning under perceptual uncertainty. 
This capability is crucial for safe and efficient robot navigation in environment with mobility-stressing elements that require terrain-specific locomotion policies.
We propose the Semantic Belief Graph (SBG), a geometric- and semantic-based representation of a robot's probabilistic roadmap in the environment. 
The SBG nodes comprise of the robot geometric state and the semantic-knowledge of the terrains in the environment. 
The SBG edges represent local semantic-based controllers that drive the robot between the nodes or invoke an information gathering action to reduce semantic belief uncertainty.
We formulate a semantic-based planning problem on SBG that produces a policy for the robot to safely navigate to the target location with minimal traversal time. 
We analyze our method in simulation and present real-world results with a legged robotic platform navigating multi-level outdoor environments.
\end{abstract}

\section{Introduction}
\begin{figure}[t]
    \centering
    \includegraphics[width=0.48\textwidth]{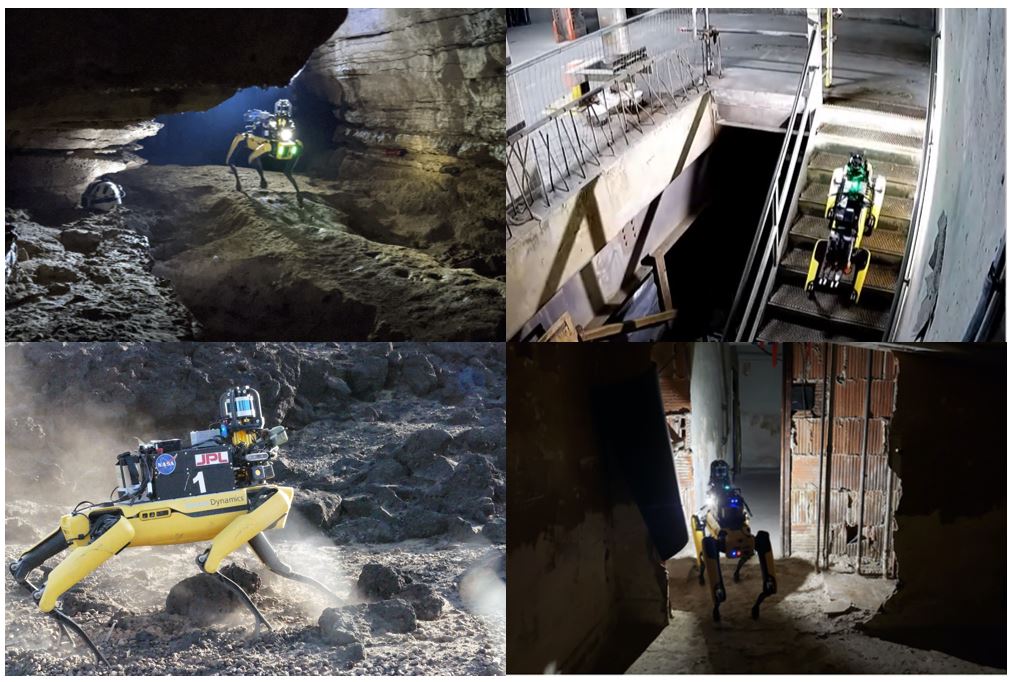}
    \caption{Boston Dynamics Spot quadrupeds endowed with NASA JPL's NeBula autonomy solution~\cite{agha2021nebula} explores challenging real-world environment by traversing various terrains including: i) slippery passages, ii) stairs, iii) rubble, and iv) narrow passages. 
    Each domain requires a specific controller to safely navigate the environment.}
    \label{fig:firstpage}
\end{figure}

Consider a robot tasked to explore complex and unknown environments autonomously. 
The environment can contain various mobility-stressing elements such as rubble, stairs, slippery surfaces, and narrow passages~(\autoref{fig:firstpage}). 
To navigate safely through these elements, the robot needs to identify the high-level semantic information of the terrain, and adapt its motion planning policy accordingly. 
However, making decisions based on semantics can be risky because of the observation noise and uncertainty of semantic recognition. 
In particular, in safety-critical systems, robot navigation using incorrect semantic classification can lead to catastrophic outcomes. 
This work investigates how to use semantic information robustly for planning in real-world operation. 

Semantic-based planning under uncertainty has three major challenges. 
The first challenge is handling the uncertainty and misclassification from the semantic detection.
The second challenge is in building a semantic representation that can be used for decision making and that is scalable to model complex and large-scale environments. 
The third challenge is in folding the semantic information into the robot planning and control framework that typically relies on geometric information.

To address these challenges, we develop a method to model semantic information for robot navigation and compute a semantic-based planning policy. 
We introduce the Semantic Belief Graph (SBG) to jointly model geometric and semantic information on the robot's probabilistic roadmap (PRM)~\cite{latombeprm}. 
The nodes in the SBG capture the uncertainty of semantic information. 
By capturing semantic uncertainty, we can model a semantic-based local controller as the SBG edges that enable the robot to choose a terrain-specific controller based on the semantic-type of the terrain. 
Moreover, we include information gathering actions in the SBG edges to model robot actions that can reduce semantic information uncertainty. 
Additionally, we discuss how to formulate the graph transition probabilities and semantic-dependent cost function. 
We show policy computation in SBG is tractable for real-world navigation problems. 

Our technical contributions are as follows:
\begin{enumerate}
    \item We propose the Semantic Belief Graph (SBG), a geo-semantic-based representation of a robot's probabilistic roadmap in the environment.
    \item By representing the semantic information in the SBG, we can predict and control the semantic uncertainty in planning rather than purely reacting to the semantic detection.
    \item We propose a new motion planning architecture that incorporates semantic information. By using high-level semantic information, this approach can perform better in navigating environments that require switching between terrain-specific controllers. 
    \item We tested the proposed method on representative simulations and in the field with a physical robot.
\end{enumerate}

\section{Related Work}
\label{sec:relworks}

\ph{Semantic Representation}
Several different approaches have been proposed for representing the high-level semantic information used for planning and navigation~\cite{crespo2020semantic}. 
Some earlier works introduce multi-hierarchical~\cite{galindo2005multi} and layered spatial representations~\cite{pronobis2012large}. 
Nuechter and Hertzberg present a semantic 3D mapping method built with onboard robotic sensors~\cite{nuchter2008towards}. 
Nieto-Granda et al. build semantic mapping by classifying regions of space~\cite{nieto2010semantic}.
There are also several works in semantic-based SLAM to estimate 3D geometry of a scene with semantic labels such as Kimera~\cite{rosinol2021kimera}, SegMap~\cite{dube2020segmap}, Semantic Fusion~\cite{mccormac2017semanticfusion}, and Voxblox++~\cite{grinvald2019volumetric}. 
In our work, we consider a graph representation to capture the semantic information and its uncertainty in the graph and associate the information directly with the robot's geometric roadmap. 
The sparsity of the graph reduces the computational and memory requirements while capturing relevant information for semantic-based planning.

\ph{Semantic-based planning for robot navigation}
Semantic-based navigation considers various planning problems and methods. 
Hahnheide et al. proposes task planning under uncertainty by classifying semantics into three knowledge layers for object search, mapping, and room categorization~\cite{hanheide2017robot}. 
Kostavelis et al. introduces a hierarchical navigation graph by building semantic maps that consist of metric, topometric, sparse topological map, and augmented navigation graphs~\cite{kostavelis2016robot}. 
Some works use learning-based methods by including semantic information in deep reinforcement learning models~\cite{yang2018visual} and imitation learning~\cite{Mousavian19}. 
In our work, in addition to using semantic knowledge for robot path planning, we also consider the problem of changing robot locomotion controllers based on semantics. 
Moreover, we also include semantic uncertainty prediction into robot actions by modeling information gathering behavior.

\ph{Semantic-based controller switching}
Our work is related to choosing suitable controller profiles based on the semantic type of the traversed terrains. 
Switching between controllers based on semantics is beneficial for efficient and safe navigation. 
One example when controller switching is critical is stair climbing because the robot needs to reliably identify stairs and perform different locomotion~\cite{helmick2002multi, mihankhah2009autonomous}. 
In legged robot navigation, switching between gait controllers is used to safely walk on slippery surfaces~\cite{dey2022prepare} and maximize multiple objectives~\cite{brandao2019multi}. 
In this work, we develop a general framework that captures the semantic uncertainty of terrain detection and reasons over which controller should be used.

\section{Problem Formulation}
\label{sec:problemformulation}
We first formulate the problem of semantic-based planning under uncertainty. 

\ph{Geo-semantic state}
We define a geo-semantic state that consists of a robot geometric state and a terrain semantic state. 
A robot geometric state $x_k \in \mathcal{X}$ is the geometric description of a robot state in the environment.
A semantic state $l_k \in \mathcal{L}$ is a type of terrain.
description in which the robot resides over. 
For example, in robot navigation, useful semantic classes include stairs, flat ground, doorways, slippery surfaces, and rubble.
The geo-semantic state is defined as a tuple of the robot and semantic state $(x_k, l_k)$

\ph{Semantic-based robot controller}
Let $u_k \in \mathcal{U}$ denote the semantic-based robot control. 
It contains controller parameters to enable the robot to traverse a specific type of terrain safely and efficiently.
The process model $(x_{k+1}, l_{k+1}) = f(x_k, l_k, u_k, w_k)$ describes how the geo-semantic state evolves as a function of the robot control and process noise $w_k$.

\ph{Observation model} In a partially observable system, the true value of the geo-semantic state is unknown and can only be inferred by the observation variable $z_k \in \mathcal{Z}$. The observation model $z_k = h(x_k, l_k, v_k)$ encodes the relation between $(x_k, l_k)$ and $z_k$, where $v_k$ is the observation noise. 

\ph{Belief} As the system operates in a partially observable setting, we use information state or belief $b_k \in \mathcal{B}$ as the data for decision making in time step $k$. We define  $b_k = p(x_k, l_k\mid z_{0:k}, u_{0:k-1})$ as a conditional probability distribution over all possible states given the history of observations and actions. Beliefs are estimated with the belief evolution model: $b_{k+1}=\tau(b_k,u^l_k,z_{k+1})$.

\ph{Policy and cost-to-go}
The policy $\pi \in \Pi$ returns the next semantic-based robot control $u_k$ given the belief $b_k$.
To determine an optimal policy $\pi^*$, we define a cost function $c(b,u)$ as a one-step cost of taking action $u$ in $b$. 
A cost function can encode different metrics to be minimized, for example, time to traverse a specific terrain using a specific controller, control energy consumption, or computational resources. 
In this formulation, the cost function calculates the traversal time.
Taking an action that risks the robot's safety, such as robot falling down stairs, will incur a high cost.

\ph{Semantic-based planning problem} We formulate the problem of semantic-based planning as follows---%
given the current robot configuration $x_S$ and semantic state $l_S$ and the goal configuration $x_G$,
find an optimal policy $\pi^*$ that moves the robot from $x_S$ to $x_G$
with minimal expected time of traverse
by selecting the best path and  semantic-based controllers: 
\begin{align}
    \label{eq:problem_statement}
    \pi^* = \argmin_{\pi \in \Pi} \mathbb{E} &\bigg{[} \sum_{k=0}^T c(b_k,u_k)    \bigg{]}\\ \nonumber
    \text{s.t.~} &b_{k+1}=\tau(b_k,u_k,z_{k+1})\\\nonumber
    &z_k = h(x_k, l_k, v_k) ,\\\nonumber
    &x_0 = x_S, l_0 = l_S, x_T = x_G , \\ \nonumber
    \forall k &\in \{0,\ldots,T\}.
\end{align}

\section{Semantic Belief Graph}
\label{sec:sbg}

In this section, we discuss the Semantic Belief Graph (SBG), a geo-semantic information roadmap in the belief space. 
The SBG is defined as a tuple of nodes and edges $(\mathcal{V}, \mathcal{M})$. 
We first introduce the formulation of SBG nodes $\mathcal{V}$ and edges $\mathcal{M}$. 
Then we discuss how to model transition probabilities and graph costs. 
Finally, we show how to find the graph policy and plan with the SBG.

\subsection{SBG nodes}
A node in SBG is a small region of belief $B^i$ around a sampled belief $b^i$. 
In the SBG graph, we parameterized the geo-semantic belief into two independent representations: a Gaussian geometric belief and a categorical semantic belief. 

\ph{Gaussian geometric belief} We assume the noise in the robot geometric belief $x_k$ is a Gaussian.
We denote the geometric random estimation vector by $x^+$, whose distribution is $p(x_k^+)=p(x_k \mid z_{0:k},u_{0:k-1})$.
The mean and covariance of $x^+$ is denoted by $\hat{x}^+ = \mathbb{E}[x^+]$ 
and $P = \mathbb{E}[(x^+ - \hat{x}^+)(x^+ - \hat{x}^+)^\intercal]$ respectively. 
In SBG nodes, the Gaussian geometric belief is characterized by the pair $(\hat{x}^+, P)$.

\ph{Categorical semantic belief}
The semantic random estimation vector $l^+$ follows a categorical distribution, whose distribution is $p(l_k^+)=p(l_k \mid z_{0:k},u_{0:k-1})$.
For a finite number of the semantic type $d_l$, $l^+$ can be represented as $(d_l+1)$-dimensional vector denoted as
$\hat{l}^+ = [l^1,l^2,\ldots,l^{d_l},l^{unknown}]$ such that $l^n \geq 0$ $\forall n$ and 
$\sum_{n=1}^{d_l+1}l^n=1$. 
The $unknown$ class is used when the semantic variable can not be classified to any of the defined semantic classes.

\ph{Constructing SBG nodes}
To construct SBG nodes $B$, we first sample PRM nodes denoted by $\{v^j\}_{j=1}^{N}$ from the obstacle-free space~\cite{kim2021plgrim}. 
We then associate the nodes with the mean and covariance of the geometric belief. 
The geometric covariance can be set according to the mission specification, for example, 
setting a small covariance to align the robot to stairs with a small tolerance.
Then we associate the semantic belief by initializing the semantic belief distribution with $unknown$ labels or with prior knowledge of the environment if any.

\subsection{SBG edges}
We design the SBG edge as a local semantic-based controller.
There are two types of semantic-based controllers: \textit{i)}~navigation controller and \textit{ii)}~information gathering controller.
A semantic-based navigation controller, denoted by $\mu_n^{ij}$ for an edge connecting nodes $B^i$ and $B^j$, drives the robot from $B^i$ to $B^j$, using a low-level motion controller designed to traverse terrain with a semantic type $l^n$.
A semantic-based information gathering controller, denoted by $\mu_{IG}^{i}$ for a self-loop edge on $B^i$, performs an information gathering action to reduce the uncertainty of the semantic classification on the current node $B^i$.

\begin{figure}[t]
    \centering
    \includegraphics[width=0.48\textwidth]{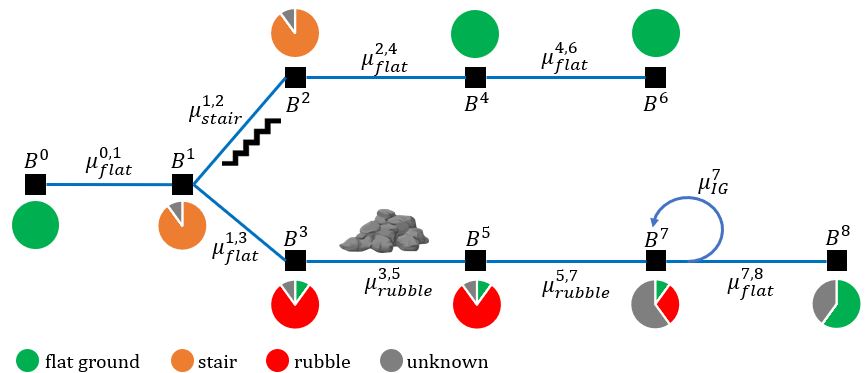}
    \caption{Visual representation of Semantic Belief Graph. 
    Graph nodes in the geo-semantic belief is denoted by $B^i$. The pie chart represents the categorical belief distribution of the semantic type around the location of the node. 
    The graph edges represent terrain specific local controller $\mu^{ij}_l$. 
    The circular edge represents information gathering action $\mu^{i}_{IG}$ to reduce semantic belief uncertainty.}
    \label{fig:sbg}
\end{figure}
\subsection{Belief Transition Probabilities}
Taking a local semantic-based controller results in updates in beliefs:
\emph{i)} the geometric belief update due to the robot configuration changes by a navigation controller $\mu_n^{ij}$, 
\emph{ii)} the semantic belief update due to additional observation by an information gathering controller $\mu_{IG}^i$.
Such belief updates, i.e., belief transition probabilities, should be modeled since the optimal policy depends on the geometric and semantic beliefs.
In this work, we assume the transition between geometric belief nodes using controller $\mu^{ij}$ is deterministic, though it is possible to extend this framework to account for failure probabilities~\cite{agha2014firm}.

To model the semantic belief transition probabilities, 
we sample a finite number of semantic belief vectors $\hat{l}^+$. 
We assume the updated semantic belief after information gathering actions will only reduce the uncertainty of the belief and reveal the most probable semantic type. 
Let $B_{k+1,j}^i$ denote a $j$-th sample of node $B^i$ at time-step $k+1$ after taking controller $\mu_{IG}^i$. Then the transition probability of the action is defined as
$P(B_{k+1,j}^i|B^i_k,\mu^{i}_{IG})$. 
The transition probability can be assumed uniform or learned from observation data.

\subsection{Graph Cost}
To find a policy on an SBG graph, we need to define the cost associated with the graph edges and nodes. 
Let $C^g(B^i,\mu)$ denote the cost of using controller $\mu$ at node $B^i$. 
The graph cost $C^g(B^i,\mu)$ sums all one-step costs along the edge until it reaches the target node $B^j$ or updated the semantic belief in current node $B^i$.

In particular, a semantic-based navigation controller $\mu^{ij}_n$ has a different cost depending on the specified mission objective and type of terrain traversed by the controller. 
In this work, we use traversal time to determine the cost. 
Using a terrain-specific controller on the wrong type of terrain incurs a higher cost as the robot locomotion will not be efficient on the wrong terrain. 
As the robot does not know the true semantic type of the terrain, $C^g(B^i,\mu^{ij}_n)$ is estimated by taking the expected value of taking controller $\mu^{ij}_n$ over the semantic distribution:
\begin{align}
    C^g(B^i,\mu^{ij}_n) = \sum_{m=1}^{n_d+1} p(l^m) C^g((x^i,l^m),\mu^{ij}_n), 
\end{align}

\subsection{Graph policy on SBG}
We define a graph policy $\pi^g:\mathcal{V}\rightarrow \mathcal{M}$ as a function that returns an edge for any given node in the graph. 
The edge is the local controller $\mu$ that then controls the robot. 
Given a target node $B^{goal}$, the graph policy $\pi^g$ can be seen as a planning tree that drives the robot to $B_{goal}$ from any node in the SBG. 

\ph{Graph cost to go}
To choose the best graph policy, we define the optimal graph-cost-to-go $J^g$ from every node. 
The cost-to-go from a given node $B^i$ is equal to the cost of the next local controller $\mu$ and the expected cost-to-go from the next node. The local controller is picked among available edges $\mathcal{M}(i)$ connected to $B^i$.
The dynamic programming (DP) equations for this graph are
\begin{align}
    J^g(B^i) = &\min_{\mu \in \mathcal{M}(i)}C^g(B^i,\mu) + 
    J^g(B^j) \\
    \pi^g(B^i)= &\argmin_{\mu \in \mathcal{M}(i)} J^g(B^i).
\end{align}
\noindent As there is a finite number of nodes in the graph, the DP equation can be solved offline. 
Standard techniques can be used to solve the DP such as with value/policy iteration methods. 

\begin{figure}[t]
    \centering
    \includegraphics[width=0.50\textwidth]{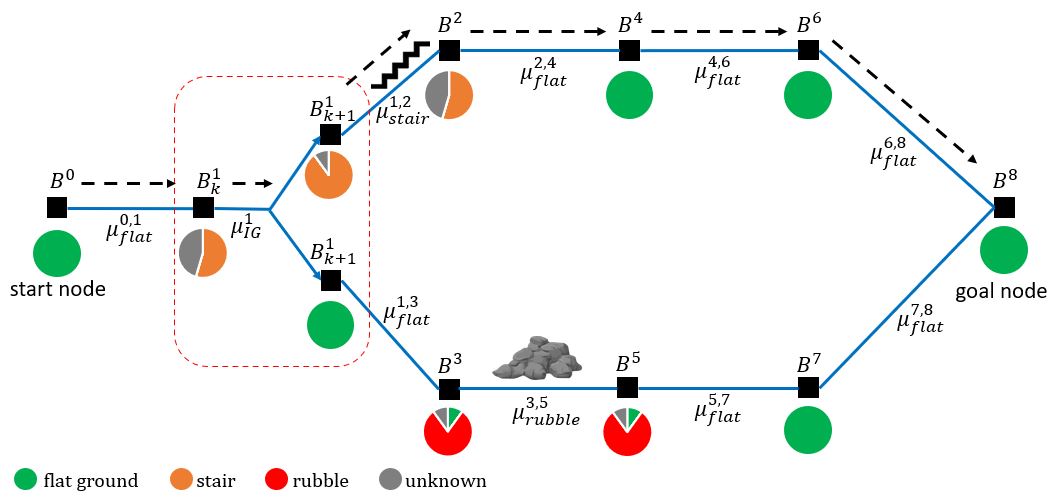}
    \caption{
    An example of a simple planning problem on SBG to move the robot from the start node to the goal node. 
    The top part of the graph shows possible path to the goal, but initially the confidence of the stair on node $B^1_k$ at time step k is low. 
    Meanwhile the bottom part of the graph shows a path through a rubble with a longer traversal time than moving through the stair. 
    The red box shows two most probable realization of the node $B^1_{k+1}$ after taking information gathering action $\mu_{IG}$. 
    The dashed arrows through the top part of the graph shows the most likely path by taking the best local controller policy on every node.
    }
    \label{fig:result1}
\end{figure}
\section{Experimental Results}
\label{sec:experiments}
We first illustrate SBG construction and show planning results in a small domain.
Then, we present simulation results in a larger environment. 
Finally, we report the results in a real-world scenario with JPL's Boston Dynamic legged robot. 

\subsection{Constructing and planning with SBG in a small environment}
In this experiment, we consider a planning problem of moving a robot to a goal in a two-level environment represented as a roadmap. 

\ph{System description}
The geo-semantic state is composed of the 3D geometric state and semantic state with four possible labels $l=\{\text{flat\_ground}, \text{stair}, \text{rubble}, \text{unknown}\}$. 
The action $u$ is generated from four possible local controllers: $\mu_{stair}$, $\mu_{flat}$, $\mu_{rubble}$, and $\mu_{IG}$. 
The first three are controllers to traverse specific types of terrain and $\mu_{IG}$ is the controller to gather semantic information around the robot. 
In this experiment, the robot is assumed to have accurate geometric localization. 
The robot observes the semantic type as a categorical distribution of possible labels and the accuracy of the semantic classification increases as the robot gets closer to the observed semantic terrain. 

\ph{SBG}
The constructed SBG and computed policy is shown in \autoref{fig:result1}. 
Every PRM node is associated with a geo-semantic belief state. 
The closed-loop graph policy is then computed with value iteration. 
We compute the optimal policy on every node in the graph. 
In the planning phase, two new nodes are sampled to account for the most possible outcome of the semantic label after the information gathering action.

\ph{Benefits of planning with SBG}
There are two key benefits by planning with SBG. 
First, by reasoning about the uncertain information regarding possible stairs in the environment, the robot can be directed to investigate a potential stair that can lead to a shorter path to the goal. 
Second, by considering the semantics of the terrain, the robot can plan to switch to different controllers that are safer and more efficient for the specific terrain type in the environment.

\subsection{Simulation Results}
\begin{figure}[t]
    \centering
    \includegraphics[width=0.48\textwidth]{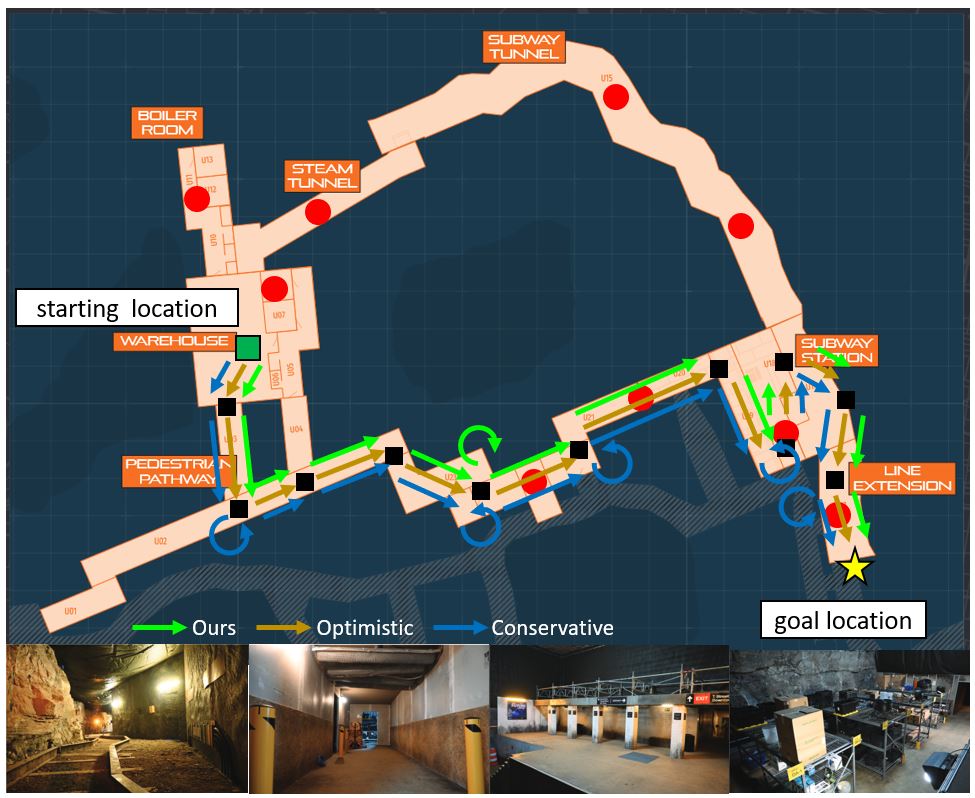}
    \caption{Course map of the simulation based on real-world urban environment depicted by the photos on the bottom. 
    Red dots on the map marked the location of stairs and rubble. The most likely paths computed by all the methods are overlaid on the map. The circular edges represent information gathering action.}
    \label{fig:map_exp2}
\end{figure}

To show that our framework is applicable to complex real-world robot navigation scenarios, we evaluate the planning performance in a simulated urban environment of the SubT Challenge Final event. 
The urban part of the course consists of 27 segments spanning around 300 m. 
The terrains and environmental model are based on the DARPA Callout document, and we use the same geo-semantic and controller model as the previous experiment. 

\begin{figure}[t]
    \centering
    \includegraphics[width=0.5\textwidth]{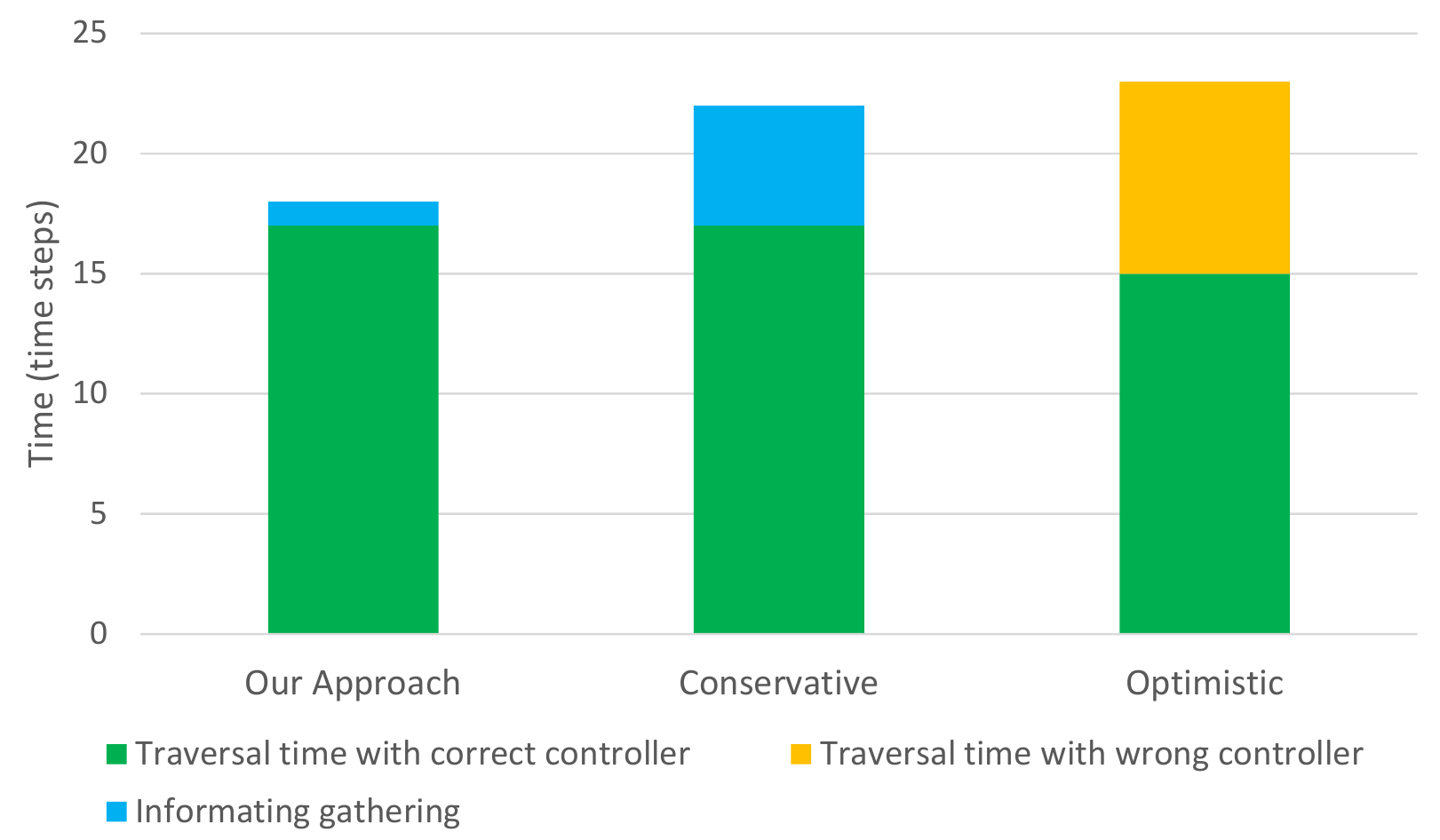}
    \caption{
    The time breakdown of the traversal time in the simulation. 
    }
    \label{fig:exp2_timeplot}
\end{figure}

We compare our method to two naive policies:
\begin{enumerate}
    \item Conservative policy: this policy only directs the robot to traverse terrains with $>95\%$ semantic confidence $l^i$, 
    \item Optimistic policy: this policy always assumes the most likely semantic type. 
\end{enumerate}
\autoref{fig:map_exp2} shows the course map of the simulation and the paths computed by all methods.
Simulation statistics are provided in \autoref{table:sim_result} and \autoref{fig:exp2_timeplot}.

\begin{table}[t]
\caption{Simulation results of the experiment in Section V.B. The traversal time (in time steps) is the average time across 20 runs taken by the robot from the starting location to the goal by following a planning policy. 
The percentage of correct terrain-specific controller is computed for all edges in the computed graph policy, including edges not on the most-likely path to the target. 
}
\centering
\begin{tabular}{@{}lrr@{}} 
 \toprule Planning policy & Percentage of correct controller & Traversal time  \\
 \midrule
 \textbf{Our Approach}  &  93.9\%  &   \textbf{18} \\
 Conservative  &  \textbf{100\%} &  22  \\
 Optimistic   &  78.8\% &  23  \\
 \bottomrule
\end{tabular}
\label{table:sim_result}
\end{table}

\ph{Traversal time}
We first compare the efficiency of the traversal time. 
The average traversal time of our method is shorter than other policies.
The conservative policy takes longer to traverse as it often performs information gathering actions to correctly identify the terrain type. 
While the optimistic policy takes longer because it uses the wrong controller for the suitable type of terrain (e.g. it uses the rubble controller to walk slowly on flat terrain). 
Our planning method on SBG balanced the two policies by accounting for the expected cost to perform information gathering or choosing an appropriate controller based on the semantic belief. 

\ph{Efficient controller switching}
We also compare the accuracy of the semantic-based controller to the true semantics of the terrain. 
While the controller accuracy of our planner is lower than the conservative policy, it balances the information gathering actions and takes a more conservative policy to minimize the overall traversal time. 
For example, in an area classified as rubble with low confidence, it saves time by choosing a more conservative policy that slows the robot movement rather than performing information gathering actions in places that takes a longer time to identify the terrain type with greater confidence.

\begin{figure}[t]
    \centering
    \includegraphics[width=0.48\textwidth]{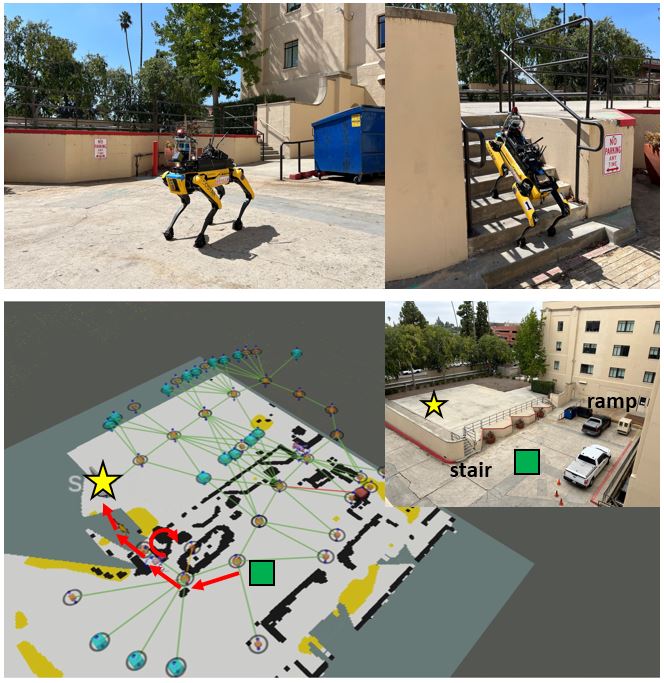}
    \caption{
    \textbf{Top}: 
    Field experiment setup in a two-level outdoor environment by using the NASA JPL Spot robot~\cite{bouman2020autonomous} and Spot's stair climbing behavior.
    \textbf{Bottom}:
    Visual representation of the policy computed with the SBG on Information Roadmaps~\cite{kim2021plgrim}. The red arrows show the most likely path from the starting location (green square) to the goal (yellow star). The circular arrow represents information gathering actions. Initially, the only path in the graph between the starting and goal location is through the ramp. By considering the path through nodes labeled with stairs, the robot chooses a shorter path by climbing the stairs.
    } 
    \label{fig:res3}
\end{figure}

\subsection{Field Test Results}
We tested our approach on the Boston Dynamics Spot legged robot in a two-level outdoor area connected by stairs and ramp as shown in \autoref{fig:res3}. 
The robot geometric localization is provided by onboard LIDAR-based SLAM solution~\cite{chang2022lamp} and the stairs are detected visually by the robot's camera and using point cloud segmentation around the robot. 
The robot has two different semantic-based controllers to walk through flat-terrain and stairs. 
The robot is assigned to move to a specific goal location on a different level. 
The prior map is built by allowing the robot to explore the environment autonomously~\cite{peltzer2022fig}. 

The computed policy to the goal on the SBG directs the robot to visit the closest stair to the goal location. 
Before attempting the stair climbing, the robot performed information gathering actions. 
The information gathering action directs the robot to pitch up to map the stair area and validate the location of the stair with the local pointcloud. 
Using probabilistic semantic information helps the robot to find a shorter path because without it the robot will take a longer route through the ramp to reach the goal. 

\section{Conclusions}
\label{sec:experiments}
We presented our method for semantic-based planning under uncertainty with SBG (Semantic Belief Graph). 
By capturing geo-semantic belief uncertainty, our method can account for it in planning and actively reduce the semantic uncertainty. 
The computed policy also selects the best terrain-based controller on every SBG edge. 
We demonstrate the benefit of our approach for time-efficient and safe motion planning in environments with a variety of challenging terrain. 
We believe the proposed method is an important step to integrate semantic knowledge for safe real-world robot navigation in such complex environments.

\vspace{-5pt}
\section*{Acknowledgment}
The work is partially supported by the Jet Propulsion Laboratory, California Institute of Technology, under a contract with the National Aeronautics and Space Administration (80NM0018D0004), and Defense Advanced Research Projects Agency (DARPA), and funded through JPL's Strategic University Research Partnerships (SURP) program.

\bibliographystyle{IEEEtran}
\bibliography{main}

\begin{thebibliography}{10}
\providecommand{\url}[1]{#1}
\csname url@samestyle\endcsname
\providecommand{\newblock}{\relax}
\providecommand{\bibinfo}[2]{#2}
\providecommand{\BIBentrySTDinterwordspacing}{\spaceskip=0pt\relax}
\providecommand{\BIBentryALTinterwordstretchfactor}{4}
\providecommand{\BIBentryALTinterwordspacing}{\spaceskip=\fontdimen2\font plus
\BIBentryALTinterwordstretchfactor\fontdimen3\font minus
  \fontdimen4\font\relax}
\providecommand{\BIBforeignlanguage}[2]{{%
\expandafter\ifx\csname l@#1\endcsname\relax
\typeout{** WARNING: IEEEtran.bst: No hyphenation pattern has been}%
\typeout{** loaded for the language `#1'. Using the pattern for}%
\typeout{** the default language instead.}%
\else
\language=\csname l@#1\endcsname
\fi
#2}}
\providecommand{\BIBdecl}{\relax}
\BIBdecl

\bibitem{agha2021nebula}
A.~Agha, K.~Otsu, B.~Morrell \emph{et~al.}, ``{NeBula: TEAM CoSTAR’s Robotic
  Autonomy Solution that Won Phase II of DARPA Subterranean Challenge},''
  \emph{Field Robotics}, vol.~2, pp. 1432--1506, 2022.

\bibitem{latombeprm}
L.~Kavraki, P.~Svestka, J.-C. Latombe, and M.~Overmars, ``Probabilistic
  roadmaps for path planning in high-dimensional configuration spaces,''
  \emph{IEEE Transactions on Robotics and Automation}, vol.~12, no.~4, pp.
  566--580, 1996.

\bibitem{crespo2020semantic}
J.~Crespo, J.~C. Castillo, O.~M. Mozos, and R.~Barber, ``Semantic information
  for robot navigation: A survey,'' \emph{Applied Sciences}, vol.~10, no.~2,
  pp. 497--524, 2020.

\bibitem{galindo2005multi}
C.~Galindo, A.~Saffiotti, S.~Coradeschi, P.~Buschka, J.-A. Fernandez-Madrigal,
  and J.~Gonz{\'a}lez, ``Multi-hierarchical semantic maps for mobile
  robotics,'' in \emph{IEEE/RSJ International Conference on Intelligent Robots
  and Systems (IROS)}, Edmonton, Canada, 2005.

\bibitem{pronobis2012large}
A.~Pronobis and P.~Jensfelt, ``Large-scale semantic mapping and reasoning with
  heterogeneous modalities,'' in \emph{IEEE International Conference on
  Robotics and Automation (ICRA)}, St. Paul, MN, 2012.

\bibitem{nuchter2008towards}
A.~N{\"u}chter and J.~Hertzberg, ``Towards semantic maps for mobile robots,''
  \emph{Robotics and Autonomous Systems}, vol.~56, no.~11, pp. 915--926, 2008.

\bibitem{nieto2010semantic}
C.~Nieto-Granda, J.~G. Rogers, A.~J. Trevor, and H.~I. Christensen, ``Semantic
  map partitioning in indoor environments using regional analysis,'' in
  \emph{IEEE/RSJ International Conference on Intelligent Robots and Systems
  (IROS)}, Taipei, Taiwan, 2010, pp. 1451--1456.

\bibitem{rosinol2021kimera}
A.~Rosinol, A.~Violette, M.~Abate, N.~Hughes, Y.~Chang, J.~Shi, A.~Gupta, and
  L.~Carlone, ``Kimera: From {SLAM} to spatial perception with 3d dynamic scene
  graphs,'' \emph{The International Journal of Robotics Research}, vol.~40, no.
  12-14, pp. 1510--1546, 2021.

\bibitem{dube2020segmap}
R.~Dube, A.~Cramariuc, D.~Dugas, H.~Sommer, M.~Dymczyk, J.~Nieto, R.~Siegwart,
  and C.~Cadena, ``Segmap: Segment-based mapping and localization using
  data-driven descriptors,'' \emph{The International Journal of Robotics
  Research}, vol.~39, no. 2-3, pp. 339--355, 2020.

\bibitem{mccormac2017semanticfusion}
J.~McCormac, A.~Handa, A.~Davison, and S.~Leutenegger, ``Semanticfusion: Dense
  3d semantic mapping with convolutional neural networks,'' in \emph{IEEE
  International Conference on Robotics and Automation (ICRA)}, Singapore, 2017.

\bibitem{grinvald2019volumetric}
M.~Grinvald, F.~Furrer, T.~Novkovic, J.~J. Chung, C.~Cadena, R.~Siegwart, and
  J.~Nieto, ``Volumetric instance-aware semantic mapping and 3d object
  discovery,'' \emph{IEEE Robotics and Automation Letters}, vol.~4, no.~3, pp.
  3037--3044, 2019.

\bibitem{hanheide2017robot}
M.~Hanheide, M.~G{\"o}belbecker, G.~S. Horn, A.~Pronobis, K.~Sj{\"o}{\"o},
  A.~Aydemir, P.~Jensfelt, C.~Gretton, R.~Dearden, M.~Janicek \emph{et~al.},
  ``Robot task planning and explanation in open and uncertain worlds,''
  \emph{Artificial Intelligence}, vol. 247, pp. 119--150, 2017.

\bibitem{kostavelis2016robot}
I.~Kostavelis, K.~Charalampous, A.~Gasteratos, and J.~K. Tsotsos, ``Robot
  navigation via spatial and temporal coherent semantic maps,''
  \emph{Engineering Applications of Artificial Intelligence}, vol.~48, pp.
  173--187, 2016.

\bibitem{yang2018visual}
W.~Yang, X.~Wang, A.~Farhadi, A.~Gupta, and R.~Mottaghi, ``Visual semantic
  navigation using scene priors,'' in \emph{International Conference on
  Learning Representations (ICLR)}, New Orleans, Louisiana, 2019.

\bibitem{Mousavian19}
A.~Mousavian, A.~Toshev, M.~Fišer, J.~Košecká, A.~Wahid, and J.~Davidson,
  ``Visual representations for semantic target driven navigation,'' in
  \emph{IEEE International Conference on Robotics and Automation (ICRA)}, 2019.

\bibitem{helmick2002multi}
D.~M. Helmick, S.~I. Roumeliotis, M.~C. McHenry, and L.~Matthies,
  ``Multi-sensor, high speed autonomous stair climbing,'' in \emph{IEEE/RSJ
  International Conference on Intelligent Robots and Systems (IROS)}, Lausanne,
  Switzerland, 2002.

\bibitem{mihankhah2009autonomous}
E.~Mihankhah, A.~Kalantari, E.~Aboosaeedan, H.~D. Taghirad, S.~Ali, and
  A.~Moosavian, ``Autonomous staircase detection and stair climbing for a
  tracked mobile robot using fuzzy controller,'' in \emph{IEEE International
  Conference on Robotics and Biomimetics}, Guangxi, China, 2009.

\bibitem{dey2022prepare}
S.~Dey, D.~Fan, R.~Schmid, A.~Dixit, K.~Otsu, T.~Touma, A.~F. Schilling, and
  A.-A. Agha-Mohammadi, ``{PrePARE}: Predictive proprioception for agile
  failure event detection in robotic exploration of extreme terrains,'' in
  \emph{IEEE/RSJ International Conference on Intelligent Robots and Systems
  (IROS)}, 2022, pp. 4338--4343.

\bibitem{brandao2019multi}
M.~Brandao, M.~Fallon, and I.~Havoutis, ``Multi-controller multi-objective
  locomotion planning for legged robots,'' in \emph{IEEE/RSJ International
  Conference on Intelligent Robots and Systems (IROS)}, Macau, China, 2019, pp.
  4714--4721.

\bibitem{kim2021plgrim}
S.~Kim, A.~Bouman, G.~Salhotra, D.~Fan, K.~Otsu, J.~Burdick, and
  A.~Agha-mohammadi, ``{PLGRIM: Hierarchical value learning for large-scale
  exploration in unknown environments},'' in \emph{International Conference on
  Automated Planning and Scheduling (ICAPS)}, Guangzhou, China, 2021.

\bibitem{agha2014firm}
A.~Agha-Mohammadi, S.~Chakravorty, and N.~M. Amato, ``Firm: Sampling-based
  feedback motion-planning under motion uncertainty and imperfect
  measurements,'' \emph{International Journal of Robotics Research}, vol.~33,
  no.~2, pp. 268--304, 2014.

\bibitem{bouman2020autonomous}
A.~{Bouman$*$}, M.~{Ginting$*$}, N.~{Alatur$*$}, M.~{Palieri}, D.~{Fan},
  T.~{Touma}, T.~{Pailevanian}, S.~{Kim}, K.~{Otsu}, J.~{Burdick}, and
  A.~Agha-mohammadi, ``Autonomous {S}pot:long-range autonomous exploration of
  extreme environments with legged locomotion,'' in \emph{IEEE/RSJ
  International Conference on Intelligent Robots and Systems (IROS)}, 2020.

\bibitem{chang2022lamp}
Y.~Chang, K.~Ebadi, C.~E. Denniston, M.~F. Ginting, A.~Rosinol, A.~Reinke,
  M.~Palieri, J.~Shi, A.~Chatterjee, B.~Morrell \emph{et~al.}, ``Lamp 2.0: A
  robust multi-robot slam system for operation in challenging large-scale
  underground environments,'' \emph{IEEE Robotics and Automation Letters},
  vol.~7, no.~4, pp. 9175--9182, 2022.

\bibitem{peltzer2022fig}
O.~Peltzer, A.~Bouman, S.-K. Kim, R.~Senanayake, J.~Ott, H.~Delecki, M.~Sobue,
  M.~J. Kochenderfer, M.~Schwager, J.~Burdick \emph{et~al.}, ``Fig-op:
  Exploring large-scale unknown environments on a fixed time budget,'' in
  \emph{IEEE/RSJ International Conference on Intelligent Robots and Systems
  (IROS)}, 2022.

\end{thebibliography}

\end{document}